\documentclass{INTERSPEECH2023}
\usepackage{multirow}
\usepackage{graphicx}
\usepackage{subcaption}
\usepackage{cite}
\usepackage{hyperref}
\hypersetup{
    colorlinks=true,
    linkcolor=red,
    urlcolor=magenta,
    }

\interspeechcameraready 

\title{Spontaneous Speech-Based Suicide Risk Detection\\ Using Whisper and Large Language Models}

\name{Z. Cui$^{1, \dagger}$, C. Lei$^{2, 3, \dagger}$, W. Wu$^{4}$, Y. Duan$^{2, 3}$, D. Qu$^{2, 3}$, J. Wu$^{1}$, R. Chen$^{2, 3, *}$, C. Zhang$^{1, *}$\thanks{$\dagger$ Co-first author. $*$ Corresponding author.}}

\address{
  {$^1$Department of Electronic Engineering, Tsinghua University, China \\
  $^2$Vanke School of Public Health, Tsinghua University, China \\
  $^3$Institute for Healthy China, Tsinghua University, China \\
  $^4$Cambridge University Engineering Department, UK}}
\email{\{cuizy20, leic22\}@mails.tsinghua.edu.cn, \{runsenchen, cz277\}@tsinghua.edu.cn}

\begin{document}

\maketitle

\begin{abstract}
The early detection of suicide risk is important since it enables the intervention to prevent potential suicide attempts. This paper studies the automatic detection of suicide risk based on spontaneous speech from adolescents, and collects a Mandarin dataset with 15 hours of suicide speech from more than a thousand adolescents aged from ten to eighteen for our experiments. To leverage the diverse acoustic and linguistic features embedded in spontaneous speech, both the Whisper speech model and textual large language models (LLMs) are used for suicide risk detection. Both all-parameter finetuning and parameter-efficient finetuning approaches are used to adapt the pre-trained models for suicide risk detection, and multiple audio-text fusion approaches are evaluated to combine the representations of Whisper and the LLM. The proposed system achieves a detection accuracy of 0.807 and an F1-score of 0.846 on the test set with 119 subjects, indicating promising potential for real suicide risk detection applications. 

\end{abstract}


\section{Introduction}

Suicide is a serious global public health issue with more than 0.7 million people dying from suicide every year around the world~\cite{world2021suicide}.
Suicide is also a leading cause of death among young individuals aged ten to twenty-four~\cite{shain2016suicide, orri2020mental}.
Early detection and intervention are crucial and serve as the most effective ways to prevent potential suicide attempts.

Diagnosis of suicide is challenging since there is no single clinical characterisation of a suicidal individual and it relies heavily on the ability, desire and honesty of a patient to communicate their symptoms, moods or cognitions~\cite{cummins2015review}.
Automatic suicide risk detection has been explored based on clinical interviews~\cite{dhelim2023artificial}, questionnaires~\cite{oh2017classification}, health records~\cite{su2020machine}, suicide notes~\cite{zhang2021automatic, ghosh2022multitask} and social contents~\cite{roy2020machine, cheng2017assessing}. 
A growing body of research has shown that speech is useful for detecting high-risk suicide individuals~\cite{cummins2015review, scherer2013investigating, stasak2021read, belouali2021acoustic}. Speech contains both semantic and non-semantic information and can be measured cheaply, remotely, non-invasively and non-intrusively. 
As an individual becomes pre-suicide, their speech undergoes discernible changes in its source features (\textit{e.g.} jitter, shimmer), prosodic features (\textit{e.g.} F0), format features, and spectral features~\cite{cummins2015review} and tends to contain more disfluency, producing more hesitations and speech errors~\cite{stasak2021read}. 
In terms of semantics, 
suicidal speech tends to have different top words from non-suicidal speech~\cite{belouali2021acoustic}. 

Foundation models are large-scale models pre-trained on vast amounts of data across a wide range of domains, which can gain specialized capabilities through finetuning or transfer learning. Foundation models have produced superior performance on many speech processing tasks such as automatic speech recognition~\cite{chang2021exploration}, emotion recognition~\cite{morais2022speech}, and detection of mental disorders and cognitive diseases such as depression~\cite{wu2023self} and Alzheimer's disease~\cite{cui2023transferring, meng2023integrated}. In particular, Large Language Models (LLM) are developing rapidly recently,
which has generated large impacts on various research fields including medical-related tasks~\cite{fan2023recommender, thirunavukarasu2023large,sadeghi2023exploring}. 
Despite its success, the use of large foundation models for automatic suicide risk detection has not been studied.

This work investigates the use of large foundation models for speech-based automatic suicide risk detection. A dataset is collected which contains spontaneous speech on suicide-related topics, due to the absence of publicly available suicide speech dataset. The dataset consists of over 1000 adolescents aged 10 to 18. Both speech foundation models and LLMs are applied to detect suicide risk. We first investigate different tuning strategies for speech and text models separately, including all-parameter finetuning (APFT) and parameter efficient finetuning (PEFT)~\cite{hu2021lora}. Then, different fusion methods are tested to combine speech and text modalities.
Furthermore, we analyze the impact of various prompt formats on LLMs' performance.
To the best of our knowledge, this is the first work that applies speech foundation models and LLMs to suicide risk detection.

The rest of the paper is organised as follows. Section~\ref{sec:dataset} describes the dataset collected in this study. Section~\ref{sec:method} introduces the proposed method for automatic suicide risk detection. The experimental setup is shown in Section~\ref{sec:exp}. Section~\ref{sec:result} presents the results and discussions. We conclude in Section~\ref{sec:conclusion}.

\section{Suicide Data}
\label{sec:dataset}

The dataset was collected from 47 primary and secondary schools in Guangdong, China.
The interviewing team consists of volunteers majoring in relevant fields such as psychology and education, who received project-specific training to ensure the professionalism of the interviews. Based on city-wide screening data, we selected at-risk students according to their history of suicide or self-harm behaviour. Additionally, we included some non-risk students as a control group for the interviews.

The interview consists of informed consent, cognitive test (including Animal Fluency Test, Digit Span Test, Short-term Memory Test and Long-term Memory Test), Mini International Neuropsychiatric Interview for Children and Adolescents (MINI-KID)~\cite{sheehan2010reliability}, as well as speech interview including question answering, reading, \textit{etc.} The interviews were conducted in quiet classrooms within the schools, utilising uniformly standardised recording devices to ensure recording quality, primarily capturing the voices of the participants. 
The ethical aspects of the project's interviews have been reviewed and approved by the Medical Ethics Committee of the Tsinghua University Technology Ethics Committee.

We collected recordings of 1179 adolescents aged 10 to 18. All recordings were conducted in Mandarin Chinese. With the result of MINI-KID, we obtained the suicide risk level of all 1179 participants, in which 631 participants (53.5\%) are at suicide risk. Detailed statistics regarding age and gender can be found in Table~\ref{table:dataset}.

In this paper, we utilised the audio recordings of participants' self-introductions, originally framed as ``How would you describe yourself?''. This question was selected due to its diverse range of responses among different participants. Beyond paralinguistic cues, these responses also contain valuable semantic information that can be effectively utilised for classification purposes. 
The total length of recordings used in this paper is 905 minutes. 
The recordings were segmented into utterances using a voice activity detection module from FunASR toolkit~\cite{gao2023funasr}. Pauses were removed between utterances. The dataset was divided into a train-dev-test split, with a ratio of 8:1:1, details shown in Table~\ref{table:dataset}. The distribution of age, gender, and suicide risk are roughly the same across the three groups.

\begin{table}[tb]
\caption{Statistics about age, gender and train-dev-test split of the collected dataset.}
\centering
\begin{tabular}{@{}ccccccc@{}}
\toprule
\multirow{2}{*}[-0.5ex]{\textbf{Age}}        & \multicolumn{2}{c}{10 to 12} & \multicolumn{2}{c}{13 to 15} & \multicolumn{2}{c}{16 to 18} \\ \cmidrule(l){2-7} 
                                    & \multicolumn{2}{c}{257}      & \multicolumn{2}{c}{546}      & \multicolumn{2}{c}{376}      \\ \midrule
\multirow{2}{*}[-0.5ex]{\textbf{Gender}}     & \multicolumn{3}{c}{\hspace{2.1em}Male}                     & \multicolumn{3}{c}{Female}                  \\ \cmidrule(l){2-7} 
                                    & \multicolumn{3}{c}{\hspace{2.1em}761}                      & \multicolumn{3}{c}{428}                     \\ \midrule
\multirow{2}{*}[-0.5ex]{\begin{tabular}[c]{@{}c@{}}\textbf{Train-Dev-Test} \\ \textbf{Split}\end{tabular}} & \multicolumn{2}{c}{Train}    & \multicolumn{2}{c}{Dev}      & \multicolumn{2}{c}{Test}     \\ \cmidrule(l){2-7} 
                                    & \multicolumn{2}{c}{944}      & \multicolumn{2}{c}{116}      & \multicolumn{2}{c}{119}      \\ \bottomrule
\end{tabular}
\vspace{-1mm}
\label{table:dataset}
\end{table}

\section{Automatic Suicide Risk Detection}
\label{sec:method}

\begin{figure}[t]
    \centering 
    \includegraphics[width=0.95\linewidth]{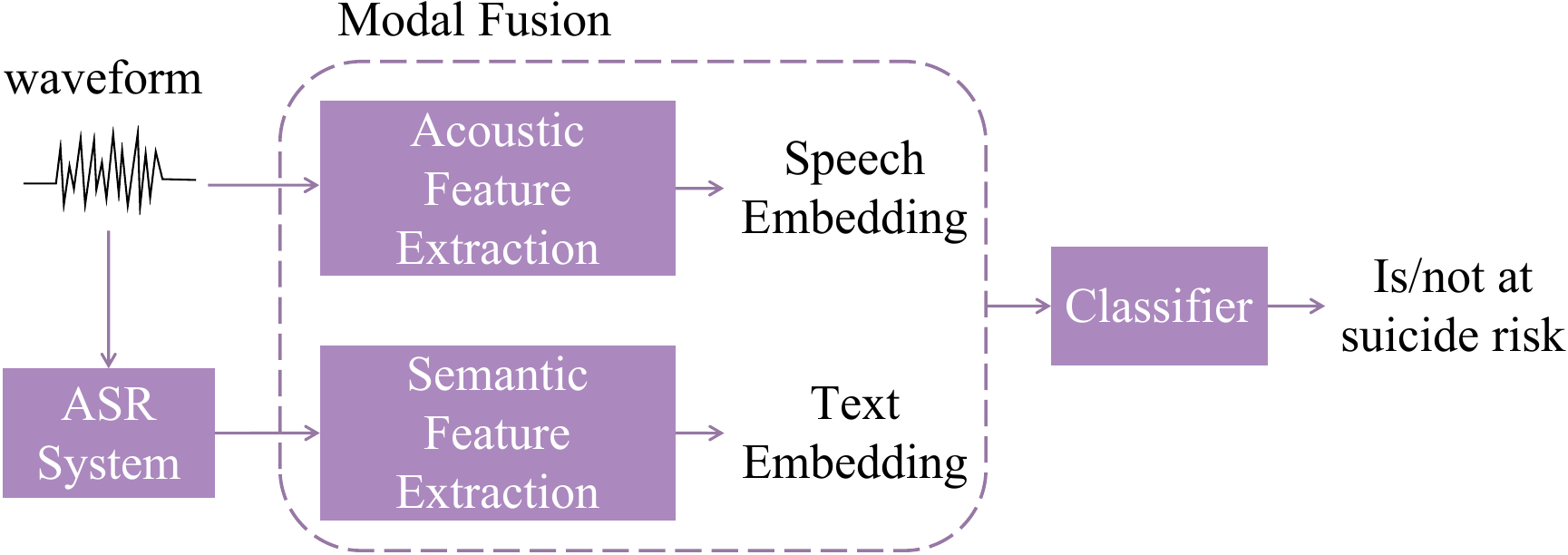}
    \caption{Overall pipeline. Modal fusion on acoustic feature and text feature is performed, followed by a classifier for detection of suicide risk.}
    \vspace{-4mm}
    \label{fig:overall}  
\end{figure}

The proposed pipeline is illustrated in Figure~\ref{fig:overall}. The system contains two branches. In the audio branch, the speech recording is fed into a speech foundation model to extract acoustic embeddings. In the text branch, an automatic speech recognition model (\textit{i.e.} a Whisper-Large model~\cite{radford2023robust})  first transcribes the input speech recording. The transcriptions are then encoded by a text foundation model to obtain text embeddings. The acoustic and text embeddings are then fused for the detection of suicide risk. Different foundation models, finetuning strategies, and fusion methods are investigated.

\subsection{Speech foundation model: Whisper}
A Whisper-Large-v3 model~\cite{radford2023robust}\footnote{https://huggingface.co/openai/whisper-large-v3} is used to extract acoustic embeddings. Whisper-Large-v3 model is a Transformer-based encoder-decoder model pre-trained on 1 million hours of weakly labelled audio and 4 million hours of pseudo-labelled audio.
The training data consists of diverse audio sources from various environments, recording conditions, speakers, and languages. 
The diversity and scale of the training data empower the Whisper model to generate high-quality audio embedding, enabling robust representation of audio content. The encoder part of the Whisper-Large-v3 model is used in this study for audio feature extraction, which contains 640M parameters. 

\subsection{Large Language Models}

For text foundation models, Baichuan2~\cite{baichuan2023baichuan2} and Qwen1.5~\cite{bai2023qwen} are compared. Both are Transformer-based LLMs. For both models, the size of 7B parameters is used in the study. The Baichuan2 LLM is open-sourced by Baichuan Intelligence Inc., which is pre-trained on a corpus with 2.6 trillion tokens. The Qwen1.5 model is proposed by Alibaba Cloud and pre-trained on 2.4 trillion tokens.

\subsection{Finetuning strategy}
Both speech and text foundation models are finetuned on the training set, which adapts the models to better recognise patterns and features relevant to identifying suicide risk.
Both APFT and PEFT are investigated. During APFT, all parameters of the foundation model are updated. For PEFT, LoRA finetuning~\cite{hu2021lora} is utilised, with weights of the pre-trained model frozen and a set of trainable rank decomposition matrices injected into each layer of the Transformer architecture, which greatly reduces the number of trainable parameters. PEFT only has 3M trainable parameters for the Whisper encoder and about 8M trainable parameters for LLM in size of 7B.

\subsection{Fusion methods}

\begin{figure}[t]
    \centering 
    \begin{subfigure}[b]{0.47\linewidth}
        \includegraphics[width=\linewidth]{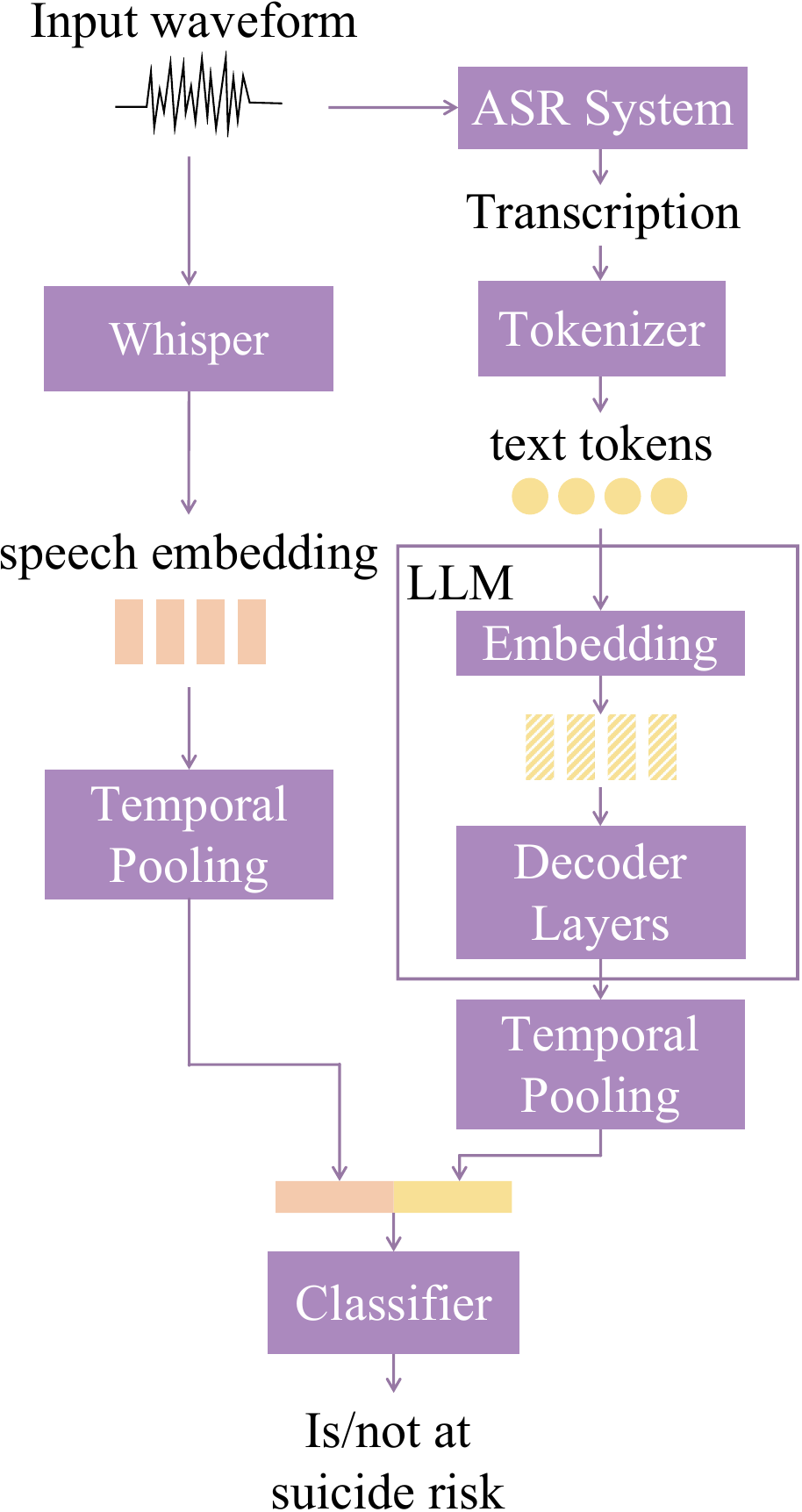}
        \caption{Concatenation fusion.}
        \label{fig:concat}
    \end{subfigure}
    \hfill
    \begin{subfigure}[b]{0.47\linewidth}
        \includegraphics[width=\linewidth]{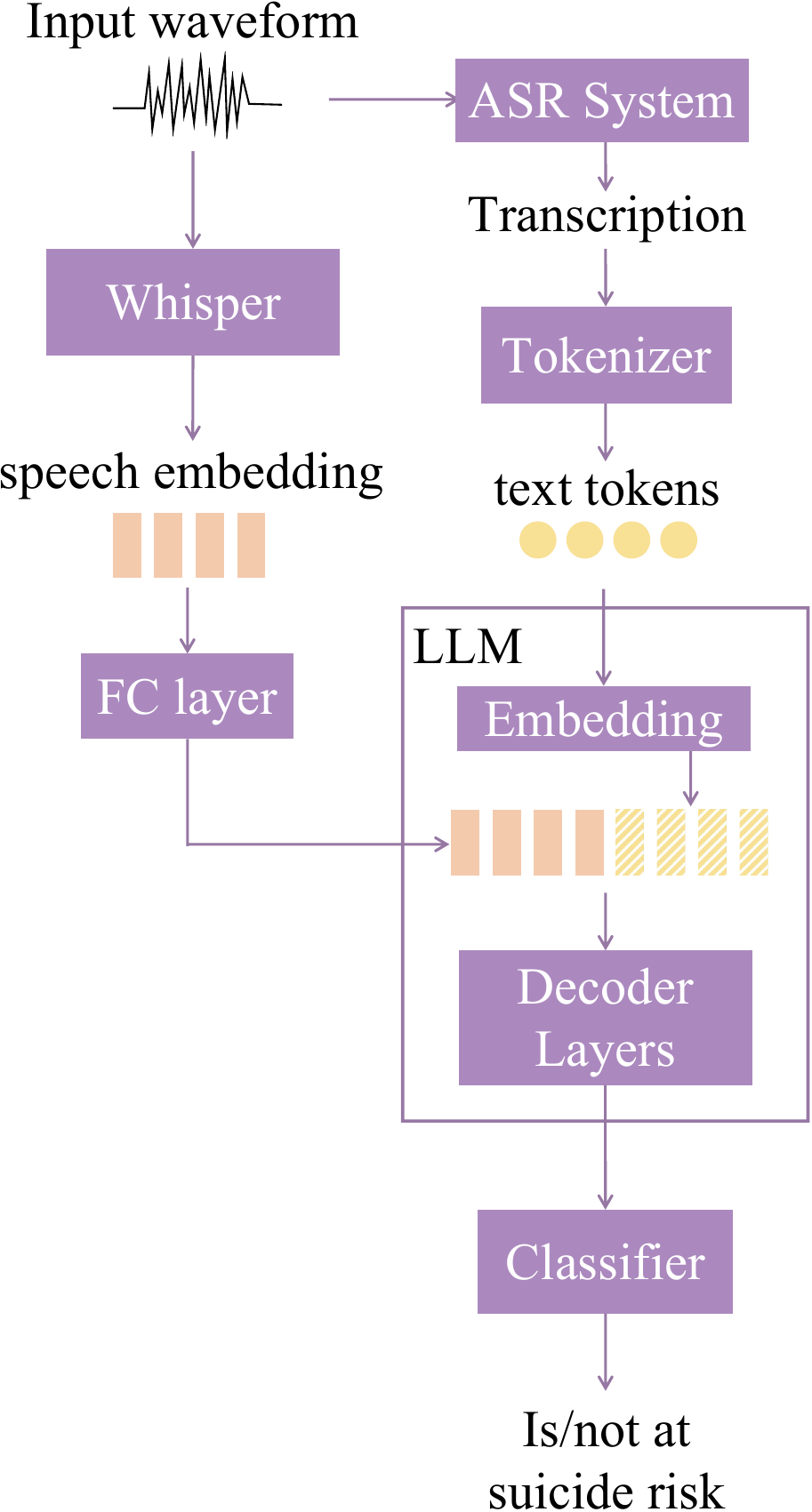}
        \caption{In-context fusion.}
        \label{fig:incontext}
    \end{subfigure}
    \caption{Model structure of two different fusion methods. For concatenation, the speech and text feature are extracted separately and pooled before concatenated. For in-context, speech embedding is mapped to LLM's hidden space and concatenated with embedded tokens, before fed into decoders of LLMs.}
    \vspace{-4mm}
    \label{fig:1}  
\end{figure}

Two types of fusion approaches are compared: concatenation fusion and in-context fusion. The detailed structure of the two types of fusion is shown in Figure~\ref{fig:concat} and Figure~\ref{fig:incontext}.

In the concatenation-based approach, temporal pooling is first applied to both acoustic embedding and text embedding. Mean pooling is adopted for the models based on the Transformer encoder (\textit{i.e.} Whisper encoder) while the last hidden state is used for models based on Transformer decoder (\textit{i.e.} LLMs) since the decoder's auto-regressive structure ensures that the final step encapsulates the most information due to its ability to leverage preceding steps for generating the current output. The pooled speech embedding and text embedding are then concatenated and fed into a fully connected layer for binary classification of suicide risk.

In the in-context fusion method, the speech embedding is fed into LLM as in-context information. 
The speech embedding first passes through a fully connected layer, mapping it from the speech model's hidden space to the LLM's hidden space. The transcriptions are tokenized and encoded by the embedding layer of the text foundation model, which is then concatenated with the mapped speech embedding and fed into the Transformer decoder layers. The classifier also consists of a fully connected layer which produces binary diagnostic results.

\section{Experiment Setup}
\label{sec:exp}

\subsection{Baselines}
The proposed method is compared to the following three baselines:
\begin{itemize}
    \item ``eGeMAPs+SVM'': a system consists of hand-crafted features and a traditional machine learning classifier. eGeMAPs feature set~\cite{eyben2015geneva} is a carefully selected collection of acoustic parameters designed to capture emotions, states, and traits in human speech. The eGeMAPS feature is extracted with the openSMILE toolkit~\cite{eyben2010opensmile}. The classification was performed using the eGeMAPS feature set as input, alongside an SVM serving as the classifier. 
    \item ``W2V2+BERT'': a system uses Wav2Vec2-XLSR-53 model~\cite{conneau2020unsupervised}\footnote{https://huggingface.co/facebook/wav2vec2-large-xlsr-53} for acoustic embeddings and  BERT-Base-Chinese model~\cite{devlin2018bert}\footnote{https://huggingface.co/google-bert/bert-base-chinese} for text embeddings, fused by concatenation.  
    \item ``Qwen-few-shot'': a 5-shot learning on Qwen1.5-7B-Chat model, which is post-trained with both supervised finetuning and direct preference optimisation to align based on Qwen1.5-7B model. 
\end{itemize}

\subsection{Implementation Details}
The system was implemented in PyTorch using the Huggingface framework.
Speed perturbation was applied when finetuning the speech foundation model. 
Sox toolkit\footnote{https://sourceforge.net/projects/sox} was used to alter the speed to 0.9 and 1.1 times the original while keeping the pitch unchanged, given the important role of pitch in suicide risk detection.
During the finetuning stage, utterances were segmented into chunks with a window length of 30 seconds and a window shift of 10 seconds. 
PEFT toolkit~\cite{peft} was utilised for LoRA finetuning.
The foundation models were finetuned for three epochs and the classifier was trained for ten epochs.
For APFT, learning rate was selected from $1\times10^{-6}$, $3 \times 10^{-6}$, and $1 \times 10^{-5}$ based on the validation set performance. For PEFT, a learning rate of $1 \times 10^{-4}$ was used. For training of the classifier, the learning rate was set to $3 \times 10^{-4}$. A linear scheduler was utilised for learning rate adjustment, with a warm-up period that covered 20\% of the total steps.

To ensure the reliability of experimental results and reduce the impact of randomness, we conducted five independent experiments for each setting using different random seeds. The average and standard deviation of these five results are reported.

\section{Results and Discussion}
\label{sec:result}

The detection results of baseline methods on the collected dataset are shown in Table~\ref{table:baseline}.
The accuracy of ``eGeMAPs+SVM''on test set is 0.537. For ``W2V2+BERT'', we got an average accuracy of 0.694 on 5 seeds. As for ``Qwen-few-shot'', 
out of 119 samples in the test set, the model failed to give responses for 45 samples. Among the 74 samples with clear response, the accuracy rate is 0.689. The overall accuracy in test set is 0.429.

\begin{table}[t]
\centering
\caption{Results of the baselines. ``eGeMAPs+SVM'' stands for eGeMAPs feature and SVM as classifier. For ``W2V2+BERT'' the average accuracy on 5 seeds is reported. The ``Qwen-few-shot'' accuracy is overall result including sample with clear response and those for which the model failed to give response.}
\begin{tabular}{@{}c|c@{}}
\toprule
\textbf{Method}           & \textbf{Accuracy} \\ \midrule
eGeMAPs+SVM          & 0.537    \\
W2V2+BERT& 0.694    \\
Qwen-few-shot  & 0.429    \\ \bottomrule
\end{tabular}
\label{table:baseline}
\end{table}

\subsection{Comparison of Different Tuning and Fusion Strategies}

\begin{table}[tb]
\centering
\caption{Results combining different tuning and fusion strategies. ``CC'' stands for concatenation fusion. ``IC'' stands for in-context fusion. The accuracy is reported in format of mean $\pm$ standard deviation.}
\resizebox{\linewidth}{!}{%
\begin{tabular}{c|cc|cc|c|cc}
\toprule
& \multicolumn{2}{c|}{\textbf{Speech}} & \multicolumn{2}{c|}{\textbf{Text}} & \multirow{2}{*}{\textbf{Fusion}} & \multirow{2}{*}{\textbf{Accuracy}} \\
 & \textbf{Model}       & \textbf{Tuning}       & \textbf{Model}      & \textbf{Tuning}      &                   &         \\                 
   \midrule
1  & Whisper        & APFT            &                &                 &                         & 0.758 $\pm$ 0.029         \\
2  & Whisper        & PEFT            &                &                 &                         & 0.633 $\pm$ 0.036         \\
3  &                &                 & Baichuan2       & APFT            &                         & 0.763 $\pm$ 0.015         \\
4  &                &                 & Baichuan2       & PEFT            &                         & 0.698 $\pm$ 0.043         \\
5  &                &                 & Qwen1.5           & APFT            &                         & 0.675 $\pm$ 0.034         \\
6  &                &                 & Qwen1.5           & PEFT            &                         & 0.737 $\pm$ 0.024         \\ \midrule
7  & Whisper        & APFT            & Baichuan2       & APFT            & CC                      & \textbf{0.768 $\pm$ 0.021}         \\
8  & Whisper        & PEFT            & Baichuan2       & PEFT            & CC                      & 0.718 $\pm$ 0.040         \\
9  & Whisper        & APFT            & Qwen1.5           & APFT            & CC                      & 0.710 $\pm$ 0.029         \\
10 & Whisper        & PEFT            & Qwen1.5           & PEFT            & CC                      & 0.752 $\pm$ 0.014         \\ \midrule
11 & Whisper        & APFT            & Qwen1.5           & APFT            & IC                      & 0.730 $\pm$ 0.033         \\
12 & Whisper        & APFT            & Qwen1.5           & PEFT            & IC                      & 0.743 $\pm$ 0.026         \\
13 & Whisper        & PEFT            & Qwen1.5           & PEFT            & IC                      & 0.694 $\pm$ 0.019         \\ \bottomrule
\end{tabular}
}
\vspace{-1mm}
\label{table:LM result}
\end{table}

The classification results of system using different tuning and fusion strategies are compared in Table~\ref{table:LM result}. As for LLMs, Baichuan2-7B-Base and Qwen1.5-7B were compared. As for tuning strategies,  APFT and PEFT were investigated. PEFT only contains less than 0.2\% trainable parameters compared with APFT. Regarding modal fusion, the use of single modal, concatenation fusion and incontext fusion were studied.

Comparing APFT and PEFT (No.~1 \& 2, No.~3 \& 4, No.~5 \& 6), it is observed that APFT produced much better results on Whisper model and Baichuan2 model, while on Qwen1.5 model PEFT got better result.

When applying concatenation fusion with APFT (No.~1, 3 \& 7), despite the good individual results of Whisper APFT and Baichuan2 APFT, a fusion of these two models yielded few improvement.
This may be attributed to the fact that both models, having undergone APFT, potentially converged to similar feature representations, thus impairing the complementarity of the features essential for effective classification.

However, when employing the concatenation fusion with PEFT (No.~2, 4 \& 8 and No.~2, 6 \& 10), it produced better performance than individually PEFT models, which indicates that speech and text models with PEFT could extract complementary information that can improve the performance of classification. It may come from its conservative alteration of the model's pre-trained attributes, due to the limited number of trainable parameters involved. But with the Baichuan2 model, though the fusion results of the PEFT model got an improvement over the individual, it's still 5\% lower than the APFT model.
It indicates that, in some cases, APFT can achieve more in-depth model optimisation, thereby obtaining higher performance, even though this may sacrifice the complementary of features. 
Furthermore, selecting an appropriate training strategy is important for different pre-trained models and downstream tasks.

Regarding the in-context fusion results, the comparison between Whisper features with APFT and PEFT (No.~12 \& 13) proves that APFT enhances the ability of the Whisper model to learn task-specific information. In addition, finetuning LLM with speech features prevents LLM from optimising towards the same feature space as the speech features. However, the worse performance on in-context fusion compared with concatenation fusion indicates that training LLMs simply with fine-tuned speech embeddings may not be sufficient and implies a need for the exploration of more powerful fusion strategies.

\begin{table}[t]
\centering
\caption{The ensemble result of our best systems. The best accuracy and F1-score of single and ensemble system is listed.}
\begin{tabular}{@{}c|ccc|c@{}}
\toprule
\textbf{System No.}     & 7     & 10    & 12    & Ensemble  \\ \midrule
\textbf{Best Accuracy} & 0.790 & 0.772 & 0.781 & \textbf{0.807} \\
\textbf{Best F1-score}  & 0.832 & 0.820 & 0.815 & \textbf{0.846} \\ \bottomrule
\end{tabular}
\vspace{-1mm}
\label{Table:vote}
\end{table}

At last, we did an ensemble of models by voting. The best accuracy and F1-score are reported in Table~\ref{Table:vote}. The index numbers refer to model numbers in Table~\ref{table:LM result}. The ensemble produced the best accuracy of 0.807 and the best F1-score of 0.846.

\begin{table}[b]
\centering
\caption{Specific prompt settings used in analysis of prompt influence. Original prompts were in Chinese and translated to English in the table.}
\label{Table:prompt setting}
\begin{tabular}{cl}
\toprule
No-prompt & Only original transcription as input.                                                                                         \\ \midrule
Prompt 1   & Determine suicide risk.                                                                                                                                                           \\ \midrule
Prompt 2   & \begin{tabular}[l]{@{}l@{}}You are a very good psychologist, the above    \\ is an interview about suicide risk, please  \\ determine whether there is a risk of suicide.\end{tabular} \\ \bottomrule
\end{tabular}
\vspace{-1mm}
\end{table}

\subsection{Analysis of Prompt Influence when Finetuning}

Previous studies have proved the efficacy of prompt engineering for the performance of LLMs~\cite{liu2023pre, qiao2023reasoning}. 
With complex instructions, LLMs achieved remarkable few/zero-shot generalizations by following these complex instructions. 
When finetuning LLMs, we assume that appropriate prompts can facilitate better optimisation towards task-specific objectives.

We examined the impact of prompts on the performance of LLMs for the suicide risk detection task. Qwen1.5-7B model and Qwen1.5-7B-Chat model, which has undergone instruction tuning, were compared. Three prompt settings were compared, as shown in Table~\ref{Table:prompt setting}.
The first is only using the original transcription as input. The second is providing a brief overview of the task information to LLM, referring to ``Prompt 1'' in the table. We also tested a detailed prompt which provides more information about the role of LLM and the description of the task, referring to ``Prompt 2'' in the table. 
Other settings are the same as No.~12 in Table~\ref{table:LM result}. 

\begin{table}[tb]
\centering
\caption{The impact of different prompt settings with LoRA finetuned Qwen1.5-7B and Qwen1.5-7B-Chat.}
\begin{tabular}{@{}cc|cc@{}}
\toprule
\textbf{Model}        & \textbf{Prompt}    & \textbf{Accuracy} \\ \midrule
Qwen1.5-7B & No-prompt & 0.703 $\pm$ 0.046   \\
Qwen1.5-7B & Prompt 1   & \textbf{0.743 $\pm$ 0.026}   \\
Qwen1.5-7B & Prompt 2   & 0.718 $\pm$ 0.044   \\
\midrule
Qwen1.5-7B-Chat & No-prompt & 0.706 $\pm$ 0.029   \\
Qwen1.5-7B-Chat & Prompt 1   & 0.727 $\pm$ 0.027   \\
Qwen1.5-7B-Chat & Prompt 2   & 0.703 $\pm$ 0.024   \\ \bottomrule
\end{tabular}
\vspace{-1mm}
\label{Table:prompt result}
\end{table}

The results are reported in Table~\ref{Table:prompt result}. Comparing different prompts of each model, Prompt 1 outperformed No-prompt. It may be because Prompt 1 provides the model with brief information about the downstream task, thereby enabling the model to better focus on task-relevant aspects, enhancing its optimisation towards the specific task. Interestingly,  Prompt 1 also outperformed Prompt 2 which provides more detailed information. 
One possible explanation is that detailed descriptions are essential for effectively guiding an LLM when it's not finetuned,  considering its general-purpose nature.
In contrast, finetuning adapts the model to the specific task, making detailed prompts less useful which reduce the information density in the input and thus impair model performance. Moreover, comparing the performance of the Base and Chat version, the Chat model exhibits about 1.5\% reduction on average accuracy. This indicates that model alignment can lead to performance reduction when finetuning LLM to downstream, even with detailed prompts.

In conclusion, this analysis suggests that when finetuning LLMs for downstream tasks, a brief prompt about the task is beneficial and sufficient. Additionally, compared to aligned models, base models perform better when finetuning is needed.

\section{Conclusions}
\label{sec:conclusion}
This paper proposes a framework for spontaneous speech-based suicide risk detection using speech and language foundation models. Both the Whisper model and LLMs were studied and different finetuning strategies and audio-text modality fusion methods were evaluated. The proposed system achieves an accuracy of 0.807 and an F1-score of 0.846, showing promising potential for real application of early suicide risk detection. 

\section{Acknowledgement}

We appreciate the efforts of Linshanjie Da, Yajing Sun, and Zeming Zhang in organizing volunteers for offline voice data collection. We are grateful to Yuhao He for his contributions to the design of the voice handbook, and to Juan Wang for her suggestions on voice handbook design and her support during the offline interview. 

\bibliographystyle{IEEEtran}
\bibliography{mybib}

\end{document}